\documentclass[conference]{ieeeconf}

\IEEEoverridecommandlockouts
\usepackage{titlesec}
\usepackage{tikz}
\usetikzlibrary{matrix}
\usepackage{tikzscale}
\usepackage{tabularray}
\usepackage{amsmath}
\usepackage{amssymb}
\usepackage{float}
\usepackage{graphicx}
\usepackage{lipsum}
\usepackage{fancyhdr}
\usepackage{siunitx}
\usepackage{comment}

\title{Collision-free time-optimal path parameterization for multi-robot teams}
\author{Katherine Mao, Igor Spasojevic, Malakhi Hopkins, M. Ani Hsieh, and Vijay Kumar %
\thanks{}
\thanks{
This works was supported by NSF Grant CCR-2112665 and the ARL DCIST CRA W911NF-17-2-0181,
Katherine Mao, Igor Spasojevic, Malakhi Hopkins, M. Ani Hsieh, and Vijay Kumar are with the GRASP Laboratory, University of Pennsylvania, PA, 19104, USA
        {\tt\small \{maokat, igorspas, mhopki3, mya, kumar\}@seas.upenn.edu}}%
        }
\begin{document}


\maketitle

\begin{abstract}
Coordinating the motion of multiple robots in cluttered environments remains a computationally challenging task. 
We study the problem of minimizing the execution time of a set of geometric paths by a team of robots with state-dependent actuation constraints. 
We propose a Time-Optimal Path Parameterization (TOPP) algorithm for multiple car-like agents, where the modulation of the timing of every robot along its assigned path is employed to ensure collision avoidance and dynamic feasibility. 
This is achieved through the use of a priority queue to determine the order of trajectory execution for each robot while taking into account all possible collisions with higher priority robots in a spatiotemporal graph.  We show a $10-20\%$ reduction in makespan against existing state-of-the-art methods and validate our approach through simulations and hardware experiments.
\end{abstract}
\section{Introduction}


In the past decade, we have seen a sharp increase in the usage of multi-robot teams. From industrial warehouse robots ferrying packages for shipment to collaborative search and rescue missions, there are numerous real-world applications that necessitate the advancement of these systems. An important component to developing these systems is the planning of collision-free, time-optimal trajectories.

This problem is challenging for several reasons.  For one, optimizing trajectories for single robots in the presence of actuation constraints can often require solving a non-convex optimization problem.  This is because actuation constraints are often state-dependent and can vary depending on a robot's position and velocity.  In general, bounds on a robot's velocity impacts how quickly a robot can move through different portions of its path, while bounds on a robot's acceleration determines its agility and ability to track a trajectory and avoid collisions with teammates.  Furthermore, even when the planning for a single robot trajectory can be posed as a convex optimization problem, the need for collision avoidance in a multi-robot setting will render it non-convex.

Second, the complexity of the planning problem scales exponentially with the number of robots in a team. This means that grid-based methods that naively discretize the joint configuration space of the team can quickly become computationally intractable despite a small team size. In fact, the problem becomes NP-hard even when only speed constraints are considered \cite{peng2005coordinating}. 
Unless P = NP, any numerical algorithm will produce either a sub-optimal solution for a limited number of problem instances or require excessive compute time.  As such, the goal is to develop an algorithmic solution that can handle both actuation and collision-avoidance constraints in a unified manner that is also computationally efficient.  

\begin{figure}[t]
    \centering
    \vspace{7pt}
    \includegraphics[width = \linewidth]{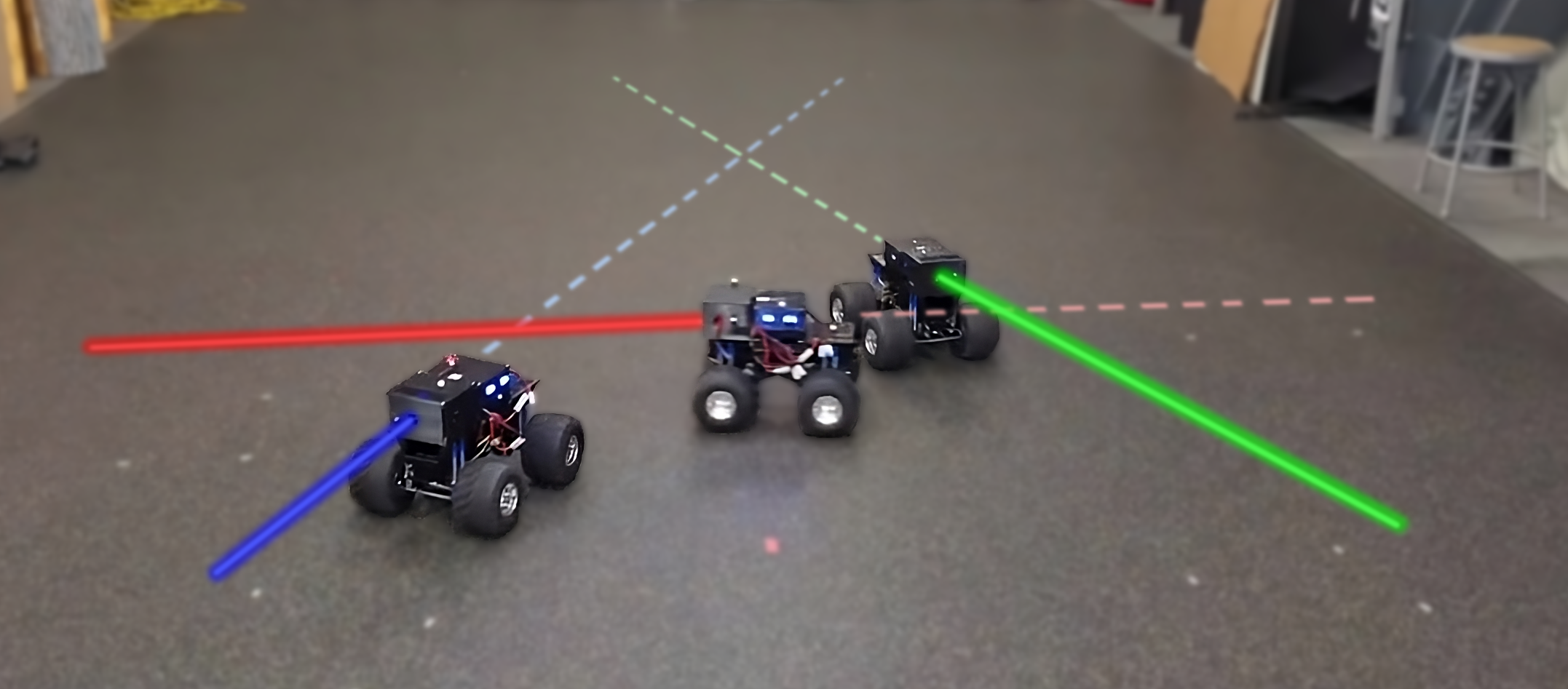}
    \caption{A set of collision-free time-optimal TOPPcar trajectories (\textbf{red}, \textbf{blue}, \textbf{green}) for a three-agent team tracked by RC cars in the a VICON motion capture space. The solid lines have already been traveled, while the dotted lines are still to be traversed.}
    \label{fig:beast_squad}
    \vspace{-10pt}
\end{figure}

\section{RELATED WORK} 
One approach for planning time-optimal trajectories is the Time-Optimal Path Parametrization (TOPP) problem \cite{bobrow1985time}. This class of problems addresses minimizing the execution time of a smooth geometric path, pre-planned to avoid static obstacles in the environment, subject to a robot's actuation constraints. 
TOPP was initially developed within the context of planning dynamically feasible trajectories for robotic manipulators \cite{bobrow1985time}. 
It has been shown to be convex for vehicles including those whose kinematics can be mapped to a bicycle model subject to constraints on acceleration, velocity, and turning rate.
For these vehicles, general convex optimization-based algorithms \cite{verscheure2009time}, as well as further refinements \cite{consolini2017optimal, pham2018new} have been proposed.

TOPP has also been extended to handle dynamic obstacles \cite{kant1986toward}. This is accomplished by decomposing the problem into a path-planning problem to avoid static obstacles and a velocity-planning problem along the resulting path to avoid dynamic obstacles. For a robot with bounds on velocity this was achieved by encoding all possible points of collision in a Path-Time ($s-t$) graph as obstacles, then iterating through all possible homotopy classes of solutions within this space. To handle bounds on acceleration, a computationally efficient algorithm for finding the optimal trajectories using a reachable velocity merging method was employed in \cite{johnson2012optimal}. Both works search for an optimal trajectory among subclasses of trajectories that alternate between extreme vertices of dynamic obstacles in the ($s-t$) \cite{kant1986toward} or Path-Velocity-Time ($s-v-t$) \cite{johnson2012optimal} graphs. These approaches rely on the assumption that feasible speeds and accelerations are independent of the robot state.
Unfortunately, it is unclear whether this assumption generalizes to state-varying actuation constraints.

For models with non-linear, non-convex actuation constraints, optimization-based approaches have been employed to directly solve the full non-convex problem \cite{liu2017speed}. Another approach iteratively linearizes the nonlinear constraints to solve the resulting convex problem, updates the candidate solution, and repeats the process until convergence \cite{xu2022speed}. 
One issue with this approach is that a different nonlinear optimization problem must be solved for every homotopy class around the dynamic obstacles in the ($s-t$) space. \cite{10285583} plans optimal trajectories among dynamic obstacles within a homotopy class of paths using non-convex optimization.

The corresponding time-optimal path parametrization problem for multi-robot systems is significantly more challenging due to the coordination of actuation constraints between robots. 
One method addresses the problem of planning multiple trajectories along fixed paths by first finding the optimal time parametrization for each agent and then optimizing the delay of the start of each robot to avoid collisions \cite{akella2002coordinating}. Follow-up work employs a Mixed-Integer Linear Programming (MILP) approach to obtain upper and lower bounds on the total execution time of the mission, or makespan, by encoding the priority order between pairs of agents around suitable over-approximations of the potential collision region \cite{peng2005coordinating}. 
However, this strategy does not take into account state-dependent actuation constraints.
There also exists decentralized approaches that optimize for both the temporal and spatial components of individual robots \cite{ma2023decentralized}.
This work iteratively treats the trajectories of other agents planned in the previous step as dynamic obstacles for agents for the current step. 
However, their heuristic approach to homotopy class selection can lead to sub-optimal trajectories.

In contrast to these existing optimization-based methods, our approach formulates the time-optimal collision-free trajectory planning problem for multiple agents in a manner that naturally includes state-dependent actuation constraints. We utilize a given priority queue to determine the order in which robot trajectories are computed. Like the previous approaches, we encode collisions between robots as obstacles in the ($s-t$) space. However, we allow the non-convex optimization solver to automatically choose the homotopy class. \textbf{The main contribution of this paper is a method for multi-agent trajectory planning with state-dependent actuation constraints.}

\section{Problem Definition}

\begin{figure}[h]
    \centering
    \includegraphics[width=0.7\linewidth]{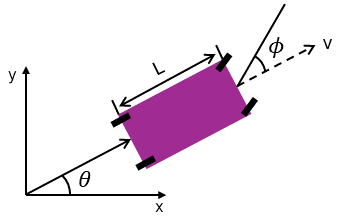}
    \caption{Diagram of the Car Model}
    \label{fig:dubins-car}
    \vspace{-10pt}
\end{figure}

We consider the problem of minimizing the makespan of traversing $N \in \mathbb{N}$ piecewise twice-continuously-differentiable regular collision-free geometric paths $\gamma^{(1)}, \dots, \gamma^{(N)}$ by a team of $N$ robots $\mathcal{R}^{(1)}, \dots, \mathcal{R}^{(N)}$.
Each robot $\mathcal{R}^{(i)}$ ($i \in [N] := \{1,\dots,N\}$) must traverse $\gamma^{(i)}$ from start to finish while respecting its own state and actuation constraints in addition to avoiding collisions with other robots in the team.
The state of $\mathcal{R}^{(i)}$ is given by
\begin{equation}
q^{(i)} = [(\textbf{p}^{(i)})^T, v^{(i)}, \theta^{(i)}]^T \in \mathbb{R}^2 \times \mathbb{R} \times S^1 =: \mathcal{X}
\end{equation}
where $\textbf{p}^{(i)} = [x^{(i)}, y^{(i)}]^T \in \mathbb{R}^2$ denotes the position of the robot, $v^{(i)} \in \mathbb{R}$ its speed, and $\theta^{(i)} \in S^1$ its orientation. 
The dynamics of $\mathcal{R}^{(i)}$ are given by 
\begin{equation}
\dot{\mathbf{q}}^{(i)} = f(\mathbf{q}^{(i)}, \mathbf{u}^{(i)}),
\end{equation}
where its control input $\mathbf{u}^{(i)} = [a^{(i)}, \phi^{(i)}]^T \in \mathbb{R} \times S^{1} =: \mathcal{U}$ consists of acceleration ($a^{(i)} \in \mathbb{R}$) and steering wheel angle ($\phi^{i} \in S^1$). 
Since derivatives with respect to different parameters will appear throughout the paper, we let $(\cdot)' = d(\cdot)/ds$ and $\dot{(\cdot)} = d(\cdot)/dt$. 
The dynamics function $f : \mathcal{X} \times \mathcal{U} \rightarrow \mathcal{X}$ is implicitly defined via 
\begin{equation}
\begin{bmatrix}
\dot{x}^{(i)} \\
\dot{y}^{(i)} \\
\dot{v}^{(i)} \\
\dot{\theta}^{(i)} \\
\end{bmatrix}
=
\begin{bmatrix}
v^{(i)}  \cos(\theta^{(i)}) \\
v^{(i)}  \sin(\theta^{(i)}) \\
a^{(i)} \\
v^{(i)} \frac{\tan(\phi^{(i)})}{L} \\
\end{bmatrix},
\end{equation}
where $L$ represents the length of the robot (see Fig \ref{fig:dubins-car}).

We impose several classes of state and actuation constraints on each $\mathcal{R}^{(i)}$. 
State constraints may consist of bounds on the translational speed of the vehicle 
\begin{equation}
0 \leq v^{(i)} \leq v_{max},
\end{equation}
as well as a bound on the magnitude of its angular velocity
\begin{equation}
| \dot{\theta}^{(i)} | \leq \omega_{max}.
\end{equation}
Actuation constraints amount to a bound on the magnitude of acceleration
\begin{equation}
||
\ddot{\textbf{p}}^{(i)}
||_{2}
\leq 
a_{max}.
\end{equation}
The path that $\mathcal{R}^{(i)}$ must traverse is given by 
\begin{equation}
\gamma^{(i)} : [0, S_{end}^{(i)}] \rightarrow \mathbb{R}^2.
\end{equation}
We assume that for all $i \in [N]$, $\gamma^{(i)}$ is a piecewise $C^2$ function and that $\frac{d}{ds}\gamma^{(i)}(s) \neq 0 \ \forall s \in [0, S_{end}^{(i)}]$. 
For the purpose of defining inter-agent collisions, we treat each $\mathcal{R}^{(i)}$ as a point mass located at $\mathbf{p}^{(i)}$.
We say that a pair of robots $\mathcal{R}^{(i)}$ and $\mathcal{R}^{(j)}$ will collide if and only if the distance between their positions is below a given minimum safe distance $d_{safe}$:
\begin{equation}
|| \textbf{p}^{(i)} - \textbf{p}^{(j)}||_2 < d_{safe},
\end{equation}
where $d_{safe}$ must be greater than or equal to the diameter of the larger robot.

Our task is to synthesize time parametrizations of $\gamma^{(1)}, ...., \gamma^{(N)}$ to minimize the maximum of their execution times (the makespan) while preventing collisions between different agents and ensuring each robot executes a dynamically feasible trajectory per the state and actuation constraints defined above. 
A time parameterization of path $\gamma^{(i)}$ is any increasing sufficiently smooth function 
\begin{equation}
\chi^{(i)} : [0, T^{(i)}] \rightarrow [0, S_{end}^{(i)}]
\end{equation} 
that satisfies $\chi^{(i)}(0) = 0, \ \chi^{(i)}(T^{(i)}) = S_{end}^{(i)}$. 
It encodes how far along the path $\mathcal{R}^{(i)}$ ought to be at each moment in time.
Since \textit{input} constraints will in general vary with the state of the robot, we define the function $\mathcal{H} : \mathcal{X} \times \mathcal{U} \rightarrow \mathbb{R}^{d_c}$ such that  $\{(x,u) \in \mathcal{X} \times \mathcal{U} \ \vert \ \mathcal{H}(x,u) \leq 0 \}$ defines the set of dynamically feasible state-input pairs. 
Summarizing everything, the problem may be stated as
\begin{equation}
\label{eq:gen_time_opt}
\begin{aligned}
& \hspace{-5mm} \min_{\substack{T^{(1)}, T^{(2)}, \dots, T^{(N)} \\ \chi^{(1)}(\cdot), \dots, \chi^{(T)}(\cdot)}} \ \max_{1 \leq i \leq N} T^{(i)} \\
\text{s.t.} & \text{ for all} \ 1 \leq i \leq N \\
& \chi^{(i)} : [0, T^{(i)}] \rightarrow [0, S_{end}^{(i)}] \ \text{is increasing} \\
& \chi^{(i)}(0) = 0, \ \chi^{(i)}(T^{(i)}) = S_{end}^{(i)} \\
& \dot{\mathbf{q}}^{(i)}(t) = f(\mathbf{q}^{(i)}(t), \mathbf{u}^{(i)}(t)) \ \forall t \in [0, T^{(i)}] \\
& \mathbf{p}^{(i)}(t) = \gamma^{(i)}(\chi^{(i)}(t)) \ \forall t \in [0, T^{(i)}] \\
& \mathcal{H}(\textbf{q}^{(i)}(t), \textbf{u}^{(i)}(t)) \leq 0 \ \forall t \in [0, T^{(i)}] \\
\text{and} & \text{ for all} \ 1 \leq i < j \leq N \\
& || \mathbf{p}^{(i)}(t) - \mathbf{p}^{(j)}(t) ||_2 \geq d_{min} \ \forall t \in [0, \max\{T^{(i)}, T^{(j)}\}].
\end{aligned}
\end{equation}

\section{Methodology}

\subsection{TOPPCar For A Single Agent}

For this work, we adapt our previous TOPPQuad formulation from \cite{mao2023toppquad} for finding time-optimal quadrotor trajectories to a car-like robot, which we refer to as TOPPCar. To streamline notation, we drop $(\cdot)^{(i)}$ where formulas apply to all $\gamma^{(i)}$.
As in \cite{verscheure2009time} the key decision variable in our optimization is the squared-speed profile $h(s) = (\frac{ds}{dt}(s))^2$. Intuitively, $h(s)$ can be thought of as a dilation factor between two trajectories that traverse the same path $\gamma^{(i)}$, modulating the time taken for an agent to reach some point $s_i$ along $\gamma^{(i)}$. Using the chain rule $\frac{d}{dt} = \frac{ds}{dt} \frac{d}{ds}$, we have the following expressions for velocity and acceleration (vectors) in terms of the square speed profile:

\begin{align}
    \label{eq:squared_speed_vel}
    & \dot{\textbf{p}} = \sqrt{h(s)}\gamma'(s), \\
    \label{eq:squared_speed_acc}
    & \ddot{\textbf{p}} = \frac{1}{2} \gamma'(s)h'(s) + \gamma''(s)h(s).
\end{align}

Taking advantage of the differentially flat nature of the car dynamics \cite{10285583}, we use $\textbf{p}$ as our flat variables to efficiently compute the states and inputs from derivatives of the geometric path $\gamma^{(i)}$ and the square speed profile. Rewriting the speed ($v$) and steering wheel angle ($\phi$) in terms of $\textbf{p}$, we get
\begin{align}
    \label{eq:diff_flat_us}
    & v = \sqrt{\dot{x}^2 + \dot{y}^2} \\
    \label{eq:diff_flat_phi}
    & \phi = \arctan{\frac{(\dot{x}\Ddot{y} - \dot{y}\Ddot{x})L}{v^3}}.
\end{align}

In addition, we can compute the tangential and normal accelerations ($a_t, a_n$) also as a function of $\textbf{p}$

\begin{align}
    \label{eq:acc_t}
    & a_t = \frac{\dot{x}\Ddot{x} + \dot{y}\Ddot{y}}{v}, \\
    \label{eq:acc_n}
    & a_n = \frac{\dot{x}\Ddot{y} - \dot{y}\Ddot{x}}{v}.
\end{align}

Using Eq. (\ref{eq:squared_speed_vel}), (\ref{eq:squared_speed_acc}), we can re-write Eq. (\ref{eq:diff_flat_us} - \ref{eq:acc_n}) as functions of $s$:

\begin{align}
    \label{eq:squared_speed_us}
    & v = \sqrt{h(s)}\sqrt{x'^2 + y'^2}, \\
    \label{eq:squared_speed_phi}
    & \phi = \arctan{\frac{(x'y'' - y'x'')L}{(x'^2 + y'^2)^{\frac{3}{2}}}}, \\
    & a_t = \frac{1}{2} h'(s)\sqrt{x'^2 + y'^2} + h(s)\frac{x'x'' + y'y''}{\sqrt{x'^2 + y'^2}},\\
    & a_n = h(s)\frac{x'y'' - y'x''}{\sqrt{x'^2 + y'^2}}.
\end{align}

A closer examination of Eq. (\ref{eq:squared_speed_phi}) shows a lack of dependence on $h(s)$, meaning its value has been fixed in the path planning stage and can be subsequently ignored in the time parameterization process. Physically, $\phi$ directs the steering angle of the front wheel and is bounded by the agent's maximum turning radius. With $\gamma^{(i)}$ fixed, it follows that this value would be independent of $h$.

We impose upper bounds on the speed of the vehicle as well as bounds on the magnitude of acceleration. 
The time optimal problem for a single robot therefore can be posed as 
\begin{equation}
\begin{aligned}
\min_{h : [0, S_{end}] \rightarrow [0, \infty)}  \int_{0}^{S_{end}} \frac{ds}{\sqrt{h(s)}} & \\
s.t. \ \forall s \in [0, S_{end}]: \hspace{18mm} & \\
|| \gamma'(s) ||_2^2 \ h(s) & \leq v_{max}^2 \\
\left|\left| \frac{1}{2} \gamma'(s)  h'(s) + \gamma''(s) h(s)  \right|\right|_2 & \leq a_{max}. \\
\end{aligned}
\label{eq:time_problem}
\end{equation}

To solve it numerically, we first discretize $[0, S_{end}]$ by a set of $M+1$ points $0 = s_0 < s_1 < \cdots < s_{M} = S_{end}$, any consecutive pair spaced $\Delta s$ apart.
Our task is to recover $(h_{k})_{k = 0}^{M}$, with $h_{k}$ being the numerical approximation of the optimal square speed profile at $s_k$ for all $0 \leq k \leq M$. 
The objective becomes 
\begin{align}
    T = \sum_{k = 0}^{M-1} \frac{2 \Delta s}{\sqrt{h_{k}} + \sqrt{h_{k+1}}}, 
\end{align} 
and the derivative of the square speed profile is approximated via forward differentiation, 
\begin{align}
    h'(s_k) = \frac{h_{k+1} - h_{k}}{\Delta s}.
\end{align}

\subsection{Collision Modeling}
\label{sec:collisions}
\begin{figure}[h]
    \centering
    \includegraphics[width=\linewidth]{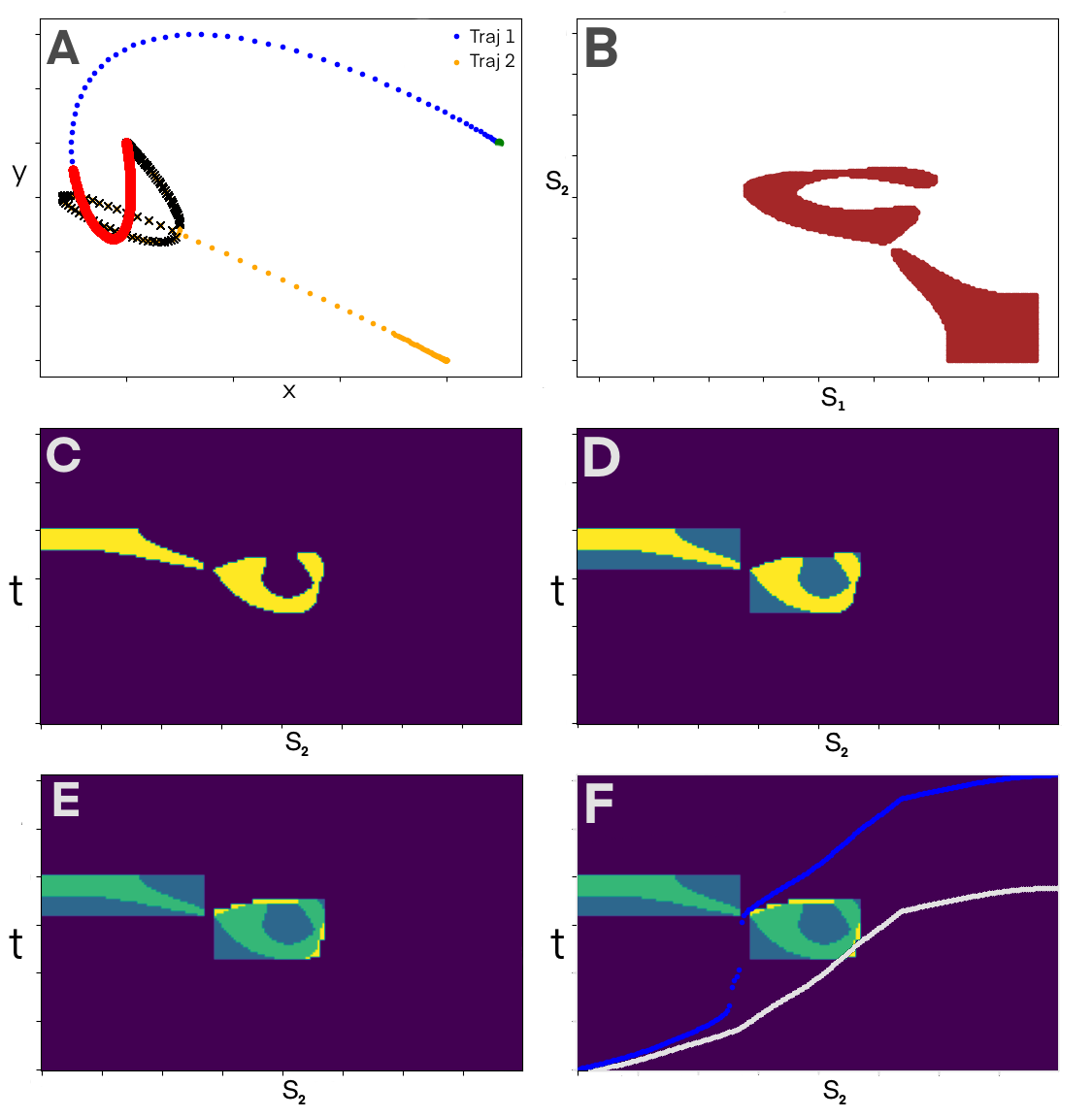}
    \caption{Depiction of the collision-avoidance constraint generation pipeline. \textbf{a)} Two discretized paths. (\textbf{blue}, \textbf{orange}) All points within a collision distance are marked (\textbf{red}. \textbf{black}) The blue trajectory has the higher priority. \textbf{b)} The points of collision plotted as collision obstacles (\textbf{red}) in ($s-s$) space. The blue trajectory is plotted along $s_1$, the horizontal axis, and the orange trajectory along $s_2$, the vertical axis. \textbf{c)} The points of collision plotted in ($s-t$) as collision obstacles (\textbf{yellow}) the orange trajectory must avoid. \textbf{d)} The unreachable spaces in ($s-t$) (\textbf{teal}), due to the monotonicity of time, completed for each obstacle. \textbf{e)} Rectangles overapproximated for each obstacle (\textbf{yellow}). \textbf{f)} TOPPcar trajectories computed with (\textbf{dark blue}) and without (\textbf{white}) collision constraints plotted in ($s-t$).} 
    \label{fig:con_pipeline}
\end{figure}

To ensure that robots avoid collisions with their teammates, we use a priority queue structure that determines the order in which we solve the TOPP problem for individual robots. 
This way, $\mathcal{R}^{(i)}$ must only guarantee collision avoidance with higher priority teammates. 
We take advantage of the discretized paths to compute a finite set of collisions for any pair of robots (Fig. \ref{fig:con_pipeline}a). 
For any pair of paths $\gamma^{(i)}, \gamma^{(j)}$, with parameters $s_{i} \in [0, S_{end}^{(i)}]$ and $s_{j} \in [0, S_{end}^{(j)}]$ respectively, potential collisions can be encoded as an occupancy map in $(s-s)$ space. 
In particular, point $(s_i, s_j)$ is deemed occupied if and only if $|| \gamma^{(i)}(s_i) - \gamma^{(j)}(s_j) || < d_{safe}$ (Fig. \ref{fig:con_pipeline}b).

At each point of collision $c$, we plot $s_{c,i}$ for $\mathcal{R}^{(i)}$'s point of collision and $t_c$, the time $\mathcal{R}^{(j)}$ traverses the same point in an ($s-t$) plot to visualize all possible points of collision spatiotemporally. This process is repeated for every robot above $\mathcal{R}^{(i)}$ in the priority queue. Spatially consecutive intersections, referred to as 'clusters', are determined from these plots (Fig. \ref{fig:con_pipeline}c). Individual clusters can be considered one group of collisions. We treat these clusters as obstacles in this ($s-t$) space that must be maneuvered around to avoid robot collisions.

We take advantage of the unique properties of the ($s-t$) graph to further reduce the possible regions a robot trajectory can occupy and to provide more structure to the collision clusters. Knowing time can only increase, any robot trajectory represented in this space must be monotonically increasing. This property allows us to complete the bottom-left and top-right corners of all ($s-t$) clusters (Fig. \ref{fig:con_pipeline}d), as those points cannot be traversed. In this process, neighboring clusters may be merged together. Additionally, we fill in any space enclosed by the clusters.

These clusters are then inscribed in a series of rectangles that overapproximate the times a collision would occur for a given obstacle (Fig. \ref{fig:con_pipeline}e). For each cluster, we compute these rectangles starting from the left-most point on the cluster and expand rightwards until the total area covered outside the cluster exceeds a specified tolerance. A maximum number of rectangles per obstacle is enforced by dynamically increasing the tolerance and re-computing until that limit is met, allowing for a trade-off between the efficiency of compute time and level of approximation conservativeness.

 To determine dynamically feasible regions of the ($s-t$) graph, we compute the TOPPCar trajectory unburdened by collision constraints and plot this trajectory in the ($s-t$) graph (Fig. \ref{fig:con_pipeline}f). All unoccupied regions that lay above this dynamic feasibility curve are reachable. For example, for a given point $s_i$, a trajectory that moves straight upwards corresponds to a robot that settles in a fixed location $\gamma^{(i)}(s_i)$ for an equivalent amount of time.

\begin{figure}[h]
    \centering
    \includegraphics[width=0.8\linewidth]{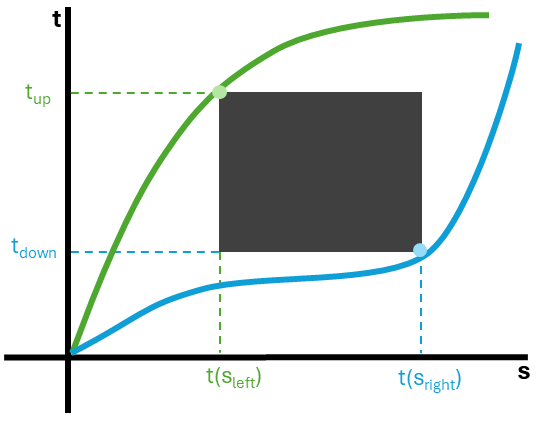}
    \caption{Illustration of time constraints due to a collision obstacle in $s-t$ with a slight abuse of notation. Two possible trajectories (\textbf{green}, \textbf{blue}) in $s-t$ space are depicted that avoid the collision obstacle (\textbf{grey}).}
    \label{fig:time_con}
\end{figure}

Finally, we use the ($s-t$) graph to add collision-avoidance constraints to Problem \ref{eq:time_problem}. For each cluster, we first determine where it lies relative to the dynamic feasibility curve. A cluster that lays below the dynamic feasibility curve is unreachable and ignored. A cluster that intersects with the dynamic feasibility curve induces, for every rectangle that covers it, a convex constraint in time, 

\begin{equation}
    t(s_{left}) > t_{up}
    \label{eq:time_con_convx}
\end{equation}
A cluster that lies fully above the dynamic feasibility curve such as Fig. \ref{fig:time_con}, induces, for every rectangle that covers it, a nonconvex constraint in time,
\begin{equation}
    (t(s_{left}) - t_{up})(t(s_{right}) - t_{down}) \geq 0
    \label{eq:time_con_nonconvex}
\end{equation}
where $t(s_{left})$, $t(s_{right})$ are the times the robot takes to reach the $s-$coordinates corresponding to the leftmost and rightmost edges of the rectangle and $t_{up}$, $t_{down}$ are the times associated with the upper and lower edges of the rectangle respectively. These constraints again arise from the monotonic nature of a trajectory plotted in ($s-t$) space. To enforce travel in the homotopy class above a rectangle, constraining $t(s_{left}) > t_{up}$ forces any resulting trajectory to remain above the rectangle. Likewise, to enforce travel in the homotopy class below, constraining $t(s_{right}) < t_{down}$ forces any resulting trajectory curve to remain below the rectangle. The structure of Eq. \eqref{eq:time_con_nonconvex} prevents the trajectory from passing through the rectangle in between $t(s_{left})$ and $t(s_{right})$.

\subsection{TOPPCar for Many Agents}

Combining the above sections, we now outline the full pipeline of our method. As stated previously, we assume a feasible priority queue, where all robots are capable of reaching their end goals without colliding into other robots or any static obstacles in the environment. For each path $\gamma^{(i)}$, we compute the corresponding TOPPCar trajectory in order of the priority queue with Problem \ref{eq:time_problem}. The first robot in the priority queue is computed without collision constraints and each subsequent robot incorporates collision constraints outlined in Sec \ref{sec:collisions} to avoid the agents that precede it. 
\section{Simulation Results}

We validate our approach in simulation against two baselines: a Fixed-Time Delay Planner \cite{akella2002coordinating} and a Decentralized Planner \cite{ma2023decentralized}. Our approach and the Fixed-Time Delay Planner are implemented in Python and we use the provided repository \cite{Car-like-Robotic-swarm} for the Decentralized Planner. All computations are performed on a laptop with an i7-6700HQ CPU. Our optimization uses the CaSADi interface to IPOPT \cite{Andersson2019} and the Fixed-Time Delay Planner uses the MILP solver in \texttt{cvxpy} \cite{diamond2016cvxpy}. Additionally, we enforce bounds of $0 \leq v \leq 5 m/s$ and $||a||_2 \leq 5 m/s^2$ in all approaches.

\subsection{Fixed-Time Delay Planner}

In the Fixed-Time Delay Planner, agent start times are delayed in a manner where all robots are able to successfully execute their trajectories without collision, with the goal of minimizing the total makespan of the robot team. We use the TOPPCar trajectory of each path $\gamma^{(i)}$ computed without the collision constraints from Sec. \ref{sec:collisions} with associated total time $T^{(i)}$ as the initial value. To improve compute time, we determine the boundary of each ($s-s$) cluster found in Sec \ref{sec:collisions} and use only those points in the MILP. This is justified because a collision at an interior cluster point forces the trajectory curve to intersect with the cluster's boundary. The full MILP formulation can be found in \cite{akella2002coordinating}.

For this comparison, we randomly generate sets of four randomized trajectories, each formed by interpolating waypoints randomly sampled from a $10m \times 10m$ box. For collision avoidance, we model the car as a sphere of radius $0.25m$. All paths are planned using a minimum jerk planner.

In Table \ref{tab:fixed_delay_baseline}, we compare the minimum increase in makespan and total trajectory time required to ensure collision avoidance and dynamic feasibility, compared to TOPPCar trajectories computed without regard for collision avoidance constraints. Over a series of 100 trials, we demonstrate that our method has, on average, a $1.12s$ shorter makespan than the Fixed-Time Delay Planner. Additionally, our method has a $2.61s$ total average delay summed over all four trajectories ($0.65s$ per trajectory) compared to the baseline's $8.77s$ total average delay ($2.19s$ per trajectory). We note that in this study our method is computed with an arbitrary feasible priority queue and not necessarily the optimal one, while the Fixed Delay baseline plans for all robots concurrently.

\begin{table}[h]
\centering
\begin{tabular}{clll}
\textbf{Approach} & \multicolumn{1}{c}{\textbf{Ours}} & \multicolumn{1}{c}{Fixed Delay} & \\ \hline
\textbf{Avg. Makespan Increase (s)}   & \textbf{1.09}     & 1.86        \\
\textbf{Avg. Total Delay (s)}        & \textbf{2.61}       & 8.77   \\
\hline
\end{tabular}
\caption{Fixed Delay Planner Baseline Comparison}
\label{tab:fixed_delay_baseline}
\vspace{-15pt}
\end{table}

\subsection{Decentralized Planner}

Additionally, we compare performance against a Decentralized Planner for multi-car trajectories. In this work \cite{ma2023decentralized}, agents attempt to reach their goals in a time-optimal manner, re-planning as they traverse their environment and encounter obstacles and each other. Although the trajectories are planned independently, this work assumes a central broadcasting network where all agents share their planned trajectories for other agents to avoid. In addition, robots have a limited sensing range to detect static obstacles.

\begin{figure}
    \centering
    \includegraphics[width=\linewidth]{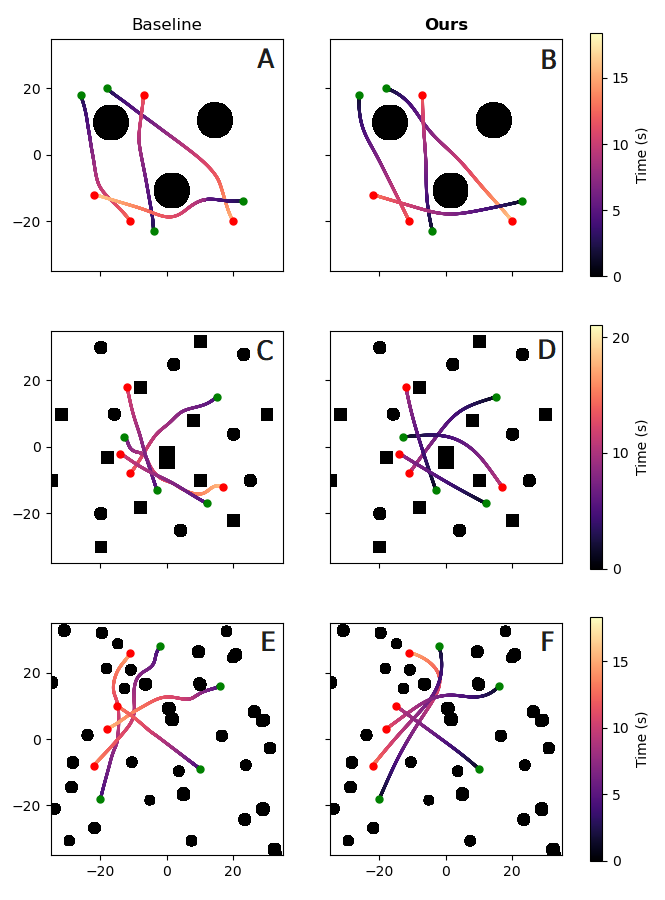}
    \caption{
    Obstacle environments for the Decentralized comparison. The starting (\textbf{green}) and ending (\textbf{red}) points of each trajectory are marked with colored dots. The paths are colorized by traversal time. Sample trajectories from three environments are depicted: \textit{A mostly-empty map} with \textbf{a)} TOPPCar trajectories and \textbf{b)} Decentralized Planner trajectories. \textit{A structured lightly-cluttered map} with \textbf{c)} TOPPCar trajectories. and \textbf{d)} Decentralized Planner trajectories. \textit{An unstructured densely-cluttered map} with \textbf{e)} TOPPCar trajectories and \textbf{f)} Decentralized Planner trajectories }
    \label{fig:decen_envs}
    \vspace{-10pt}
\end{figure}

We compare the average performance of ten sets of trajectories for four agents across three environments: a mostly-empty map  (Fig. \ref{fig:decen_envs}a,b), a structured lightly-cluttered map (Fig. \ref{fig:decen_envs}c,d), and an unstructured more densely-cluttered map (Fig. \ref{fig:decen_envs}e,f). For each trajectory, start and goal waypoints that satisfy the collision-free assumption are randomly selected from a $70m \times 70m$ box. To align with the provided car model in \cite{Car-like-Robotic-swarm}, we model the robot as a sphere of radius $1.75
m$. All paths are planned with a minimum-jerk planner. To make the comparison as fair as possible, we increase the sensing range of the Decentralized Planner baseline to encompass the map of the whole environment. 

\begin{table}[h]
\centering
\begin{tabular}{clll}
\textbf{Environment} & \multicolumn{1}{c}{Avg.$\Delta$ Makespan (s)} & \multicolumn{1}{c}{Avg. $\Delta$  Traj (s)} & \\ \hline
\textbf{Env 1}   & 1.57    & 1.57       \\
\textbf{Env 2}       & 3.21       & 2.99   \\
\textbf{Env 3}       & 1.69       & 1.58   \\
\hline
\end{tabular}
\caption{Decentralized Planner Baseline Comparison}
\label{tab:fixed_horizon_baseline}
\vspace{-15pt}
\end{table}

Because our planner can utilize a different path than the Decentralized Planner, we compare the difference in makespan and total trajectory time between the two approaches.
From Table \ref{tab:fixed_horizon_baseline}, we show a $1.5-3s$ decrease in both average makespan and average trajectory execution times compared to the Decentralized Planner, resulting in a $10-20\%$ trajectory traversal time reduction. This can mainly be credited to our planner's awareness of the spatiotemporal collisions as trajectories are planned. In the Decentralized Planner, robots must frequently decelerate and accelerate as they encounter and re-plan around one another during execution, which can result in significant delays. Our approach takes these collision into account at the planning stage, which allows robots to maintain a higher average velocity. Additionally, for \cite{ma2023decentralized} to run online, certain assumptions must be made on how homotopy classes between obstacles are chosen. During the re-planning phase, these assumptions can occasionally result in large detours which substantially add to the total trajectory execution time.

There are situations where the ability to re-plan trajectories can be beneficial. One such situation arises when the optimal path, ignorant of other agents, funnels all agents between the same set of obstacles (Fig. \ref{fig:decen_envs}f). Intuitively, it would be beneficial in minimizing makespan if some robots were to deviate their paths around an adjacent obstacle. With our algorithm, the total makespan for this trial was $18.76s$, with the last robot having to wait until the first robot has almost completed its full trajectory, while the makespan of the Decentralized Planner is $13.30s$. Overall though, we have shown that our planner offers significant time advantages for planning in environments with known static obstacles, while the ability to re-plan trajectories can be beneficial in spaces with unknown obstacles.

\section{Hardware Demonstration}

\begin{figure}[h]
    \centering
    \includegraphics[width=\linewidth]{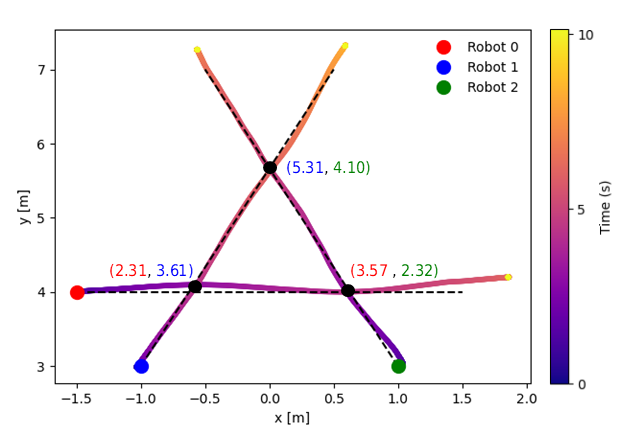}
    \caption{The planned (\textbf{dotted}) and tracked (\textbf{solid}) trajectories of a team of three RC Cars following collision-free time-optimal TOPPCar trajectories at $1m/s$. The starting point of each robot is marked with a dot of its associated color (\textbf{red}, \textbf{green}, \textbf{blue}). The points of collision are marked (\textbf{black}) and labeled with the times of arrival color-coded to the corresponding robot. The plotted paths are colorized by traversal time.}
    \label{fig:hardware_demo}
    \vspace{-10pt}
\end{figure}

We demonstrate the performance of our algorithm with a hardware deployment on a team of RC cars (Fig. \ref{fig:beast_squad}). Each vehicle weighs $8.16kg$, has a $0.35m \times 0.42m$  footprint, and a turning radius of of $1.5m$. The robots are $4\times4\times4$ wheel drive, with two motors and two servos that control the front and back wheel pairs separately. For safety, we limit the maximum velocity of each vehicle to $1m/s$. The robots are powered by a Raspberry Pi 4 and controlled through ROS Noetic using a proportional controller to send waypoint and velocity commands. Experiments are performed in a VICON motion capture system to collect ground truth state information. All trajectories are computed off-board, with commands sent from a computer base station at 1000Hz. As seen in Fig. \ref{fig:hardware_demo}, all three agents are capable of tracking their respective trajectories, reaching a maximum velocity of $1m/s$ without colliding into each other.

\section{Conclusion}

In this paper, we present a collision-free time-optimal path parameterization trajectory planning approach for multiple robots. 
Our method takes as input pre-determined paths that avoid static obstacles.
Then, inter-robot collisions are determined by a priority queue and plotted in a spatiotemporal graph as collision obstacles. 
Our method then uses a non-convex optimization approach to re-parameterize the temporal traversal of each path to ensure collision avoidance between robots.
We compare the performance of our planner against other state-of-the-art time-optimal multi-agent trajectory planners and show a $10-20\%$ reduction in makespan.
Of course, with priority queue assumption comes the inherent lack of optimality guarantees due to limiting the set of considered homotopy classes between obstacles.
Future work will explore how best to optimize these priority queues and explore potential solutions that do not require them.

\bibliographystyle{IEEEtran}
\bibliography{references_new}

\begin{thebibliography}{10}
\providecommand{\url}[1]{#1}
\csname url@samestyle\endcsname
\providecommand{\newblock}{\relax}
\providecommand{\bibinfo}[2]{#2}
\providecommand{\BIBentrySTDinterwordspacing}{\spaceskip=0pt\relax}
\providecommand{\BIBentryALTinterwordstretchfactor}{4}
\providecommand{\BIBentryALTinterwordspacing}{\spaceskip=\fontdimen2\font plus
\BIBentryALTinterwordstretchfactor\fontdimen3\font minus \fontdimen4\font\relax}
\providecommand{\BIBforeignlanguage}[2]{{%
\expandafter\ifx\csname l@#1\endcsname\relax
\typeout{** WARNING: IEEEtran.bst: No hyphenation pattern has been}%
\typeout{** loaded for the language `#1'. Using the pattern for}%
\typeout{** the default language instead.}%
\else
\language=\csname l@#1\endcsname
\fi
#2}}
\providecommand{\BIBdecl}{\relax}
\BIBdecl

\bibitem{peng2005coordinating}
J.~Peng and S.~Akella, ``Coordinating multiple robots with kinodynamic constraints along specified paths,'' \emph{The international journal of robotics research}, vol.~24, no.~4, pp. 295--310, 2005.

\bibitem{bobrow1985time}
J.~E. Bobrow, S.~Dubowsky, and J.~S. Gibson, ``Time-optimal control of robotic manipulators along specified paths,'' \emph{The international journal of robotics research}, vol.~4, no.~3, pp. 3--17, 1985.

\bibitem{verscheure2009time}
D.~Verscheure, B.~Demeulenaere, J.~Swevers, J.~De~Schutter, and M.~Diehl, ``Time-optimal path tracking for robots: A convex optimization approach,'' \emph{IEEE Transactions on Automatic Control}, vol.~54, no.~10, pp. 2318--2327, 2009.

\bibitem{consolini2017optimal}
L.~Consolini, M.~Locatelli, A.~Minari, and A.~Piazzi, ``An optimal complexity algorithm for minimum-time velocity planning,'' \emph{Systems \& Control Letters}, vol. 103, pp. 50--57, 2017.

\bibitem{pham2018new}
H.~Pham and Q.-C. Pham, ``A new approach to time-optimal path parameterization based on reachability analysis,'' \emph{IEEE Transactions on Robotics}, vol.~34, no.~3, pp. 645--659, 2018.

\bibitem{kant1986toward}
K.~Kant and S.~W. Zucker, ``Toward efficient trajectory planning: The path-velocity decomposition,'' \emph{The international journal of robotics research}, vol.~5, no.~3, pp. 72--89, 1986.

\bibitem{johnson2012optimal}
J.~Johnson and K.~Hauser, ``Optimal acceleration-bounded trajectory planning in dynamic environments along a specified path,'' in \emph{2012 IEEE International Conference on Robotics and Automation}.\hskip 1em plus 0.5em minus 0.4em\relax IEEE, 2012, pp. 2035--2041.

\bibitem{liu2017speed}
C.~Liu, W.~Zhan, and M.~Tomizuka, ``Speed profile planning in dynamic environments via temporal optimization,'' in \emph{2017 IEEE Intelligent Vehicles Symposium (IV)}.\hskip 1em plus 0.5em minus 0.4em\relax IEEE, 2017, pp. 154--159.

\bibitem{xu2022speed}
W.~Xu and J.~M. Dolan, ``Speed planning in dynamic environments over a fixed path for autonomous vehicles,'' in \emph{2022 International Conference on Robotics and Automation (ICRA)}.\hskip 1em plus 0.5em minus 0.4em\relax IEEE, 2022, pp. 3321--3327.

\bibitem{10285583}
Z.~Han, Y.~Wu, T.~Li, L.~Zhang, L.~Pei, L.~Xu, C.~Li, C.~Ma, C.~Xu, S.~Shen, and F.~Gao, ``An efficient spatial-temporal trajectory planner for autonomous vehicles in unstructured environments,'' \emph{IEEE Transactions on Intelligent Transportation Systems}, vol.~25, no.~2, pp. 1797--1814, 2024.

\bibitem{akella2002coordinating}
S.~Akella and S.~Hutchinson, ``Coordinating the motions of multiple robots with specified trajectories,'' in \emph{Proceedings 2002 IEEE International Conference on Robotics and Automation (Cat. No. 02CH37292)}, vol.~1.\hskip 1em plus 0.5em minus 0.4em\relax IEEE, 2002, pp. 624--631.

\bibitem{ma2023decentralized}
C.~Ma, Z.~Han, T.~Zhang, J.~Wang, L.~Xu, C.~Li, C.~Xu, and F.~Gao, ``Decentralized planning for car-like robotic swarm in cluttered environments,'' in \emph{2023 IEEE/RSJ International Conference on Intelligent Robots and Systems (IROS)}.\hskip 1em plus 0.5em minus 0.4em\relax IEEE, 2023, pp. 9293--9300.

\bibitem{mao2023toppquad}
K.~Mao, I.~Spasojevic, M.~A. Hsieh, and V.~Kumar, ``Toppquad: Dynamically-feasible time optimal path parametrization for quadrotors,'' \emph{arXiv preprint arXiv:2309.11637}, 2023.

\bibitem{Car-like-Robotic-swarm}
\BIBentryALTinterwordspacing
C.~Ma, Z.~Han, T.~Zhang, J.~Wang, L.~Xu, C.~Li, C.~Xu, and F.~Gao. Decentralized car-like robotic swarm. [Online]. Available: \url{https://github.com/ZJU-FAST-Lab/Car-like-Robotic-swarm.git}
\BIBentrySTDinterwordspacing

\bibitem{Andersson2019}
J.~A.~E. Andersson, J.~Gillis, G.~Horn, J.~B. Rawlings, and M.~Diehl, ``{CasADi} -- {A} software framework for nonlinear optimization and optimal control,'' \emph{Mathematical Programming Computation}, vol.~11, no.~1, pp. 1--36, 2019.

\bibitem{diamond2016cvxpy}
S.~Diamond and S.~Boyd, ``{CVXPY}: {A} {P}ython-embedded modeling language for convex optimization,'' \emph{Journal of Machine Learning Research}, vol.~17, no.~83, pp. 1--5, 2016.

\end{thebibliography}

\end{document}